\pdfoutput=1
\documentclass[11pt]{article}

\PassOptionsToPackage{hyphens}{url}
\usepackage[final]{acl}

\usepackage{times}
\usepackage{latexsym}
\usepackage{amsmath}   % provides \text{} for subscripts in math

\usepackage[T1]{fontenc}
\usepackage[utf8]{inputenc}

\usepackage{microtype}

\usepackage{inconsolata}

\usepackage{graphicx}

\usepackage{multirow}
\usepackage{booktabs}
\usepackage{tcolorbox}

\usepackage{array}
\usepackage{ragged2e}

\usepackage{xcolor}

\usepackage[capitalize]{cleveref}
\crefname{section}{Sec.}{Secs.}
\Crefname{section}{Section}{Sections}
\crefname{appendix}{Appendix}{Appendices}
\Crefname{appendix}{Appendix}{Appendices}
\Crefname{table}{Table}{Tables}
\crefname{table}{Tab.}{Tabs.}
\Crefname{figure}{Figure}{Figures}
\crefname{figure}{Fig.}{Figs.}

\usepackage{xspace}
\makeatletter
\DeclareRobustCommand\onedot{\futurelet\@let@token\@onedot}
\def\@onedot{\ifx\@let@token.\else.\null\fi\xspace}

\def\eg{\emph{e.g}\onedot}

\makeatother

\title{Beyond the Context Window: A Cost-Performance Analysis of Fact-Based Memory vs. Long-Context LLMs for Persistent Agents}

\author{Natchanon Pollertlam$^{1}$ \quad Witchayut Kornsuwannawit$^{1}$\\
$^{1}$ Bricks Technology \\
{\tt  $^{1}$ \{natchanon.p,witchayut.k\}@brickstech.co}
}

\begin{document}
\maketitle

\begin{abstract}

Persistent conversational AI systems face a choice between passing full conversation histories to a long-context large language model (LLM) and maintaining a dedicated memory system that extracts and retrieves structured facts.
We compare a fact-based memory system built on the Mem0 framework against long-context LLM inference on three memory-centric benchmarks—LongMemEval, LoCoMo, and PersonaMem~v2—and evaluate both architectures on accuracy and cumulative API cost.
Long-context GPT-5-mini achieves higher factual recall on LongMemEval and LoCoMo, while the memory system is competitive on PersonaMem~v2, where persona consistency depends on stable, factual attributes suited to flat-typed extraction.
We construct a cost model that incorporates prompt caching and show that the two architectures have structurally different cost profiles: long-context inference incurs a per-turn charge that grows with context length even under caching, while the memory system's per-turn read cost remains roughly fixed after a one-time write phase.
At a context length of 100k tokens, the memory system becomes cheaper after approximately ten interaction turns, with the break-even point decreasing as context length grows.
These results characterize the accuracy–cost trade-off between the two approaches and provide a concrete criterion for selecting between them in production deployments.

\end{abstract}
\section*{Executive Summary}

When people use an AI assistant over many days or months — for customer support, personal help, or learning — the AI needs to remember what was said in earlier conversations. There are two main ways to do this. The first approach keeps a full record of every past conversation and sends the whole record to the AI every time a new question is asked. The second approach saves only the key facts from past conversations into a small memory store, and pulls out only what is needed when a question comes in. Both approaches work, but they have different strengths and costs. This paper compares the two to help organizations decide which one is right for their situation.

We tested both approaches using three separate tests that measure how well an AI assistant remembers things about a user over time — such as facts, dates, and personal details. We also tracked how much each approach costs as users ask more and more questions, taking into account price discounts that AI service providers offer when the same conversation is sent repeatedly, which is called a \textbf{prompt caching discount}.

On accuracy, the full-history approach was more correct on two of the three tests, by a margin of about 33 to 35 percentage points. On the third test — which checks whether the AI stays consistent about a user's personal details, such as their habits and preferences — the memory-based approach performed just as well and was even slightly better in one comparison. This means that the full-history approach is not always the winner; it depends on what kind of questions are being asked.

On cost, the picture changes as users interact more. In the first few interactions, the full-history approach costs less. However, because it sends the entire conversation history every time, its cost keeps growing with each new question. The memory-based approach pays a small one-time cost upfront to build the memory store, and then charges only a small, steady fee for each new question. When a conversation history reaches about 100,000 words, the memory-based approach becomes cheaper after around 10 questions, and saves about 26\% of the total cost by the time 20 questions have been asked. The longer the conversation history grows, the sooner this saving comes into effect.

In short, there is no single best approach for every situation. For services where users come back many times — such as ongoing customer support, personal assistants, or tutoring tools — the memory-based approach saves a meaningful amount of money while still giving good answers. For one-time or short interactions where accuracy is the top priority, the full-history approach is the better choice. This paper gives organizations a simple way to calculate which approach will cost less, based on how many times their users are likely to interact with the AI.

\section{Introduction}
\label{sec:intro}

As conversational AI assistants are deployed in settings that span multiple sessions---personal assistants, customer-support agents, educational tutors---they must maintain coherent access to information from past interactions. Two strategies address this requirement. The first passes the raw conversation history directly into the context window of a long-context large language model (LLM), relying on the model to process all prior turns on every query. The second builds a dedicated memory system that distills interactions into compact, structured records and retrieves only the relevant facts at query time~\cite{chhikara2025mem0buildingproductionreadyai,maharana2024evaluatinglongtermconversationalmemory}. As context windows have expanded to millions of tokens~\cite{team2025gemini25,singh2025openaigpt5card}, some have questioned whether retrieval-based approaches remain necessary~\cite{khalusova2024ragvslc,rengifo2025longercontext}. However, the practical case for memory systems rests not only on accuracy but on the economics of long-running deployments.

API pricing for long-context LLMs scales with input length, and re-submitting a 100k-token history on every user turn accumulates cost rapidly. Prompt caching partially alleviates this by discounting tokens that share a common prefix with a recently processed prompt~\cite{gim2024promptcachemodularattention}---OpenAI charges cached input tokens at 10\% of the standard rate for GPT-5-series models, and Anthropic applies the same 90\% discount across its Claude models~\cite{openai2024promptcaching,openai2025pricing,anthropic2024promptcaching}---but each turn still incurs an incremental charge proportional to context length. Memory systems have a structurally different cost profile: a one-time write phase covers fact extraction and embedding, while each subsequent query consumes only the retrieved facts, which are far smaller than the full history. Whether the memory approach is economically preferable depends on how frequently a user re-engages with the same context---a quantity we refer to as the \emph{break-even point}.

In this paper, we compare a fact-based memory system built on the Mem0 framework~\cite{chhikara2025mem0buildingproductionreadyai} against long-context LLM inference on three memory-centric benchmarks: LongMemEval, LoCoMo~\cite{maharana2024evaluatinglongtermconversationalmemory}, and PersonaMem v2. In addition to measuring factual recall accuracy, we construct a cost model that incorporates prompt caching and tracks cumulative API expenditure as a function of the number of interaction turns. Our main contributions are:

\begin{itemize}
    \item An accuracy comparison across three benchmarks of varying context length and question type, showing that long-context LLMs achieve higher factual recall on most datasets, while the memory system remains competitive on persona-consistency tasks.
    \item A cost model incorporating prompt caching that demonstrates the memory system's fixed per-turn read cost is structurally different from the cached but still growing per-turn cost of long-context inference.
    \item A break-even analysis showing that at a context length of 100k tokens the memory system becomes cheaper after approximately ten interaction turns, and that this threshold decreases as context length grows.
\end{itemize}

\section{Related Work}
\label{sec:related}

\noindent
\textbf{Long-context LLMs.}
Extending the effective context window of transformer-based models has been studied through architectural changes~\cite{guo2022longt5,beltagy2020longformer} and positional encoding techniques~\cite{chen2023extending}. Recent models such as Gemini 2.5 Pro~\cite{team2025gemini25}, Claude Opus 4.5~\cite{anthropic2025claude45opus}, and GPT-5~\cite{singh2025openaigpt5card} support context windows of one million tokens or more, making it possible in principle to load full conversation histories into a single prompt. In practice, however, performance often degrades when relevant information appears in the middle of a long context~\cite{liu2024lost}, and effective context utilization has been shown to be substantially shorter than the nominal window size~\cite{hsieh2024ruler}. Additional input length can also impair reasoning in some settings~\cite{levy2024same}. To reduce the computational cost of long-context inference, prior work has proposed prompt compression~\cite{jiang2023longllmlingua} and model cascading~\cite{chen2023frugalgpt}. In this work, we treat a long-context LLM with prompt caching as one of two candidate architectures for persistent conversational memory, and we evaluate its accuracy and cost against a fact-based memory system.

\noindent
\textbf{Memory systems for conversational agents.}
Several types of memory have been identified for LLM agents: working memory covers the active context window; factual (long-term) memory persists user-specific information such as preferences and past events; and episodic memory records event-ordered sequences of interactions~\cite{packer2024memgptllmsoperatingsystems,su2026dialoguetimetemporalsemantic}. Retrieval-augmented generation~\cite{lewis2020retrieval} is an early form of memory augmentation, prepending retrieved document chunks to the prompt to ground generation in external knowledge. MemGPT~\cite{packer2024memgptllmsoperatingsystems} draws an analogy between LLM context management and OS virtual memory, introducing a tiered storage hierarchy that pages facts between in-context working memory and external long-term storage. More recently, Mem0~\cite{chhikara2025mem0buildingproductionreadyai} introduces a pipeline that extracts atomic, flat-typed facts from conversation history and stores them in a vector database; at query time, the top-$k$ facts are retrieved as context for the answer model. This design separates the extraction cost (a one-time write phase) from the query cost (a fixed read phase), and achieves a 91\% reduction in p95 response latency compared to full-context processing on the LoCoMo benchmark~\cite{maharana2024evaluatinglongtermconversationalmemory,chhikara2025mem0buildingproductionreadyai}. More recent systems pursue richer representations. EverMemOS~\cite{hu2026evermemosselforganizingmemoryoperating} organizes facts into hierarchical memory scenes through an engram-inspired lifecycle of episodic trace formation, semantic consolidation, and reconstructive recollection. \citet{su2026dialoguetimetemporalsemantic} propose Temporal Semantic Memory (TSM), which distinguishes between the time a conversation is recorded and the real-world time events occur, and consolidates temporally continuous facts into durative summaries that capture persistent user states. A-MEM~\cite{xu2025amemagenticmemoryllm} takes an agentic approach inspired by the Zettelkasten method, constructing memory notes with LLM-generated contextual attributes and autonomously establishing semantic links between related memories; new experiences can further trigger updates to existing memory representations. Zep~\cite{rasmussen2025zeptemporalknowledgegraph} structures agent memory as a temporally-aware knowledge graph that tracks historical relationships between entities and maintains fact validity periods, enabling reasoning about how entity states evolve across sessions. These systems represent a progression from flat fact stores toward structured, temporally grounded representations. Our work uses Mem0's flat-typed extraction as a controlled baseline and compares it against long-context inference rather than against other memory architectures.

\noindent
\textbf{Cost efficiency in LLM inference.}
The per-token pricing of LLM APIs makes cost a central concern in production deployments. Prompt caching---reusing the precomputed key-value states of recently seen input prefixes~\cite{gim2024promptcachemodularattention}---is among the most effective cost-reduction mechanisms: OpenAI charges cached input tokens at 10\% of the standard rate for GPT-5-series models, and Anthropic applies the same 90\% discount across its Claude models~\cite{openai2024promptcaching,openai2025pricing,anthropic2024promptcaching}. Cache-Augmented Generation~\cite{Chan_2025} applies caching to static knowledge bases, loading documents into the KV cache as an alternative to retrieval. \citet{zhang2025tailoptimizedcachingllminference} analyze KV cache eviction policies and show that the standard Least Recently Used policy can be suboptimal for tail latency in multi-turn conversations, motivating more principled cache management strategies. In our cost model, prompt caching is applied to the long-context baseline at a 90\% discount on input tokens from the second turn onward, consistent with the GPT-5-series pricing used in our experiments~\cite{openai2024promptcaching,openai2025pricing}; we then derive the number of turns at which the memory system becomes cheaper.

\noindent
\textbf{Long-context benchmarks and evaluation.}
Several benchmarks evaluate LLMs on tasks that require memory of extended interactions. LongMemEval~\cite{wu2025longmemevalbenchmarkingchatassistants} tests five core abilities---information extraction, multi-session reasoning, temporal reasoning, knowledge updates, and abstention---with conversation histories of approximately 115k tokens per question in the standard single-session setting. LoCoMo~\cite{maharana2024evaluatinglongtermconversationalmemory} provides ten multi-session dialogues with nearly 2,000 questions spanning single-hop, multi-hop, temporal, and open-domain categories. PersonaMem v2~\cite{jiang2025personamemv2personalizedintelligencelearning} evaluates whether a system maintains consistent responses about user-specific persona information across a series of questions. Broader long-context benchmarks not specific to conversational memory include LongBench~\cite{bai2023longbench}, $\infty$Bench~\cite{zhang2024infty}, and the Needle-in-a-Haystack community benchmark~\cite{greg2023needle}. Studies using these benchmarks have documented the lost-in-the-middle degradation pattern~\cite{liu2024lost}, discrepancies between nominal and effective context size~\cite{hsieh2024ruler}, and challenges in retrieval from very long sequences~\cite{kuratov2024searchneedles11mhaystack}.

\section{Methodology}
\label{sec:method}

We conducted a comparative study to evaluate the performance and economic feasibility of fact-based memory systems against long-context LLMs. This section outlines the architectural setup of our memory baseline, the datasets employed, the evaluation metrics, and the framework for our cost-performance analysis.

\subsection{Memory System Baseline (Mem0)}
\label{sec:mem0-baseline}

The memory-augmented baseline was implemented using the Mem0 Open Source framework~\cite{chhikara2025mem0buildingproductionreadyai} with a custom configuration. The system architecture consists of four primary stages. Full prompt templates are provided in \Cref{sec:prompts}.

\textbf{Conversation Segmentation:} Input dialogues are processed in segments using \texttt{batch\_size=10}, with a maximum content length of 8,000 characters per segment. By processing dialogues in chronological windows, segmentation preserves temporal order and prevents context drift during fact extraction.

\textbf{Fact Extraction:} GPT-5-nano~\cite{singh2025openaigpt5card} served as the fact extractor model, configured with low reasoning effort and default API parameters (see \Cref{sec:model-params}). Long-form conversations are distilled into atomic, non-hierarchical (flat-typed) facts to ensure compatibility with vector-based retrieval. This process converts raw context into structured memory records.

\textbf{Embedding \& Storage:} Extracted facts are embedded into a 1536-dimensional vector space using the text-embedding-3-small model.\footnote{\url{https://developers.openai.com/api/docs/models/text-embedding-3-small}} These are stored in a pgvector\footnote{\url{https://github.com/pgvector/pgvector}} vector database utilizing an HNSW index (\texttt{m=16}, \texttt{ef\_construction=64}, cosine similarity).

\textbf{Retrieval Mechanism:} For each query, the system performs a standard vector search with \texttt{top\_k=20} to retrieve relevant facts. The reasoner model (GPT-5-mini~\cite{singh2025openaigpt5card}) consumes the retrieved context, using default API parameters (see \Cref{sec:model-params}).

\subsection{Long-context LLM}
\label{sec:long-context-llm}

Aside from the memory system baseline, we evaluate a long-context LLM that uses no memory extraction. Instead of extracting and indexing facts, the raw conversation history is passed directly to the answer-generation LLM as retrieved context. The experimental setup uses GPT-5-mini~\cite{singh2025openaigpt5card} and GPT-OSS-120B~\cite{openai2025gptoss120bgptoss20bmodel} for answer generation (default API parameters; see \Cref{sec:model-params}), with timestamps included in the conversation format.

\subsection{Experimental Datasets}
\label{sec:datasets}
% Use table* and figure* for full-width floats to avoid two-column overlap.

We evaluated the systems across three benchmarks representing different context lengths and memory requirements. \Cref{tab:dataset-conv} and \Cref{tab:dataset-q} summarize the conversation and question token statistics per dataset (counted with \texttt{o200k\_base} tokenizer~\cite{tiktoken}).

\begin{table*}[t]
\centering
\begin{tabular}{l r r r r r}
\toprule
\textbf{Dataset} & $N$ & Min & Max & Med. & Mean \\
\midrule
LongMemEval & 500 & 95,271 & 103,935 & 101,747 & 101,601 \\
LoCoMo      &  10 &  9,690 &  19,242 &  17,788 &  15,966 \\
PersonaMem v2 & 222 & 14,828 &  28,479 &  23,546 &  22,997 \\
\bottomrule
\end{tabular}
\caption{Conversation token statistics per dataset.}
\label{tab:dataset-conv}
\end{table*}

\begin{table*}[t]
\centering
\begin{tabular}{l r r r r r}
\toprule
\textbf{Dataset} & $N$ & Min & Max & Med. & Mean \\
\midrule
LongMemEval & 500 & 5 & 65 & 15 & 17.8 \\
LoCoMo      & 1,986 & 4 & 28 & 11 & 12.0 \\
PersonaMem v2 & 589 & 7 & 163 & 28 & 58.4 \\
\bottomrule
\end{tabular}
\caption{Question token statistics per dataset.}
\label{tab:dataset-q}
\end{table*}

\textbf{LongMemEval}~\cite{wu2025longmemevalbenchmarkingchatassistants} tests the limits of context windows and retrieval precision with conversations exceeding 100k tokens. \textbf{LoCoMo}~\cite{maharana2024evaluatinglongtermconversationalmemory} focuses on complex reasoning over long histories with many questions per conversation. \textbf{PersonaMem v2}~\cite{jiang2025personamemv2personalizedintelligencelearning} evaluates the system's ability to maintain consistent agentic personas.

\subsection{Evaluation Metrics: LLM-as-a-Judge}
\label{sec:eval-metrics}

To assess the accuracy of generated responses, we utilized an LLM-as-a-judge framework.

\textbf{Judge Model:} GPT-5-mini~\cite{singh2025openaigpt5card} was selected as the evaluator due to its high reasoning capabilities and long-context support.

\textbf{Consensus Protocol:} To mitigate stochastic variance and ``self-preference'' bias~\cite{zheng2023judgingllmasajudgemtbenchchatbot}, we implemented a 3-vote consensus (Majority Vote) system. For every question, the judge model independently evaluated the alignment between the generated answer and the ``Golden Answer'' three times; the final score was determined by the majority outcome.

\textbf{Accuracy Calculation:} Accuracy is defined as the percentage of questions where the generated response was judged as correct relative to the total number of questions in the dataset.

\subsection{Cost-Performance Analysis Framework}
\label{sec:cost-analysis}

The economic evaluation compares the cumulative cost of the Memory System ($C_{\text{Mem}}$) against the Long-context LLM ($C_{\text{LC}}$) over $N$ interaction turns (where $N \in \{1, 5, 10, 20\}$).

\subsubsection{Cost Equations}
\label{sec:cost-equations}

The cost models incorporate Prompt Caching (providing a 90\% discount on input tokens for $N > 1$\footnote{\label{fn:pricing}\url{https://openai.com/api/pricing/}}) and are defined as follows:

\begin{enumerate}
\item \textbf{Long-Context LLM Cost} ($C_{\text{LC}}$):
\begin{equation}
\begin{split}
C_{\text{LC}}(N) &= C_{\text{first\_turn}} \\
&\quad + (N-1) \times C_{\text{cached\_turn}}
\end{split}
\end{equation}
Where $C_{\text{first\_turn}}$ reflects the full price of the 100k+ context, and $C_{\text{cached\_turn}}$ reflects the discounted rate for subsequent queries using the same context.

\item \textbf{Memory System Cost} ($C_{\text{Mem}}$):
\begin{equation}
C_{\text{Mem}}(N) = C_{\text{write}} + N \times C_{\text{read}}
\end{equation}
$C_{\text{write}}$: A one-time initialization cost for fact extraction and embedding (using GPT-5-nano and text-embedding-3-small pricing\textsuperscript{\ref{fn:pricing}}). $C_{\text{read}}$: The turn-by-turn cost of processing only the retrieved facts and the current query (using GPT-5-mini pricing).
\end{enumerate}

\subsubsection{Break-even Analysis}
\label{sec:break-even}

We calculated the Break-even Point ($N_{\text{BE}}$), the turn count at which $C_{\text{Mem}} < C_{\text{LC}}$. This analysis was performed across varying context lengths ($L$) to determine the Sensitivity of Economic Scalability in relation to data volume.

\subsection{Experimental Environment}

All LLM and embedding API calls were routed through OpenRouter.\footnote{\url{https://openrouter.ai}} The vector database used pgvector running on PostgreSQL~17.\footnote{\url{https://github.com/pgvector/pgvector}} Full prompt templates for fact extraction and LLM-as-a-judge evaluation are provided in \Cref{sec:prompts}.

\section{Experiments and Results}
\label{sec:experiments}

We evaluate the memory system and the long-context baselines along two dimensions: \textbf{accuracy} on factual recall tasks and \textbf{cost} as a function of interaction volume. All models and pricing details are described in \Cref{sec:mem0-baseline,sec:long-context-llm,sec:cost-analysis}.

% ── 4.1 Accuracy ─────────────────────────────────────────────────────────────

\subsection{Accuracy Results}
\label{sec:accuracy}

\Cref{tab:accuracy-main} reports accuracy across the three evaluation datasets for the Memory System (Mem0), LC GPT-5-mini, and LC GPT-OSS-120B.

\begin{table*}[ht]
\centering
\begin{tabular}{l r r r}
\toprule
\textbf{Dataset} & \textbf{Memory System} & \textbf{LC GPT-5-mini} & \textbf{LC GPT-OSS-120B} \\
\midrule
LoCoMo           & 57.68 & \textbf{92.85} & 81.69 \\
PersonaMem v2    & 62.48 & \textbf{69.75} & 60.50 \\
LongMemEval      & 49.00 & \textbf{82.40} & 48.20 \\
\bottomrule
\end{tabular}
\caption{Accuracy (\%) on three datasets~\cite{maharana2024evaluatinglongtermconversationalmemory,jiang2025personamemv2personalizedintelligencelearning,wu2025longmemevalbenchmarkingchatassistants}. ``Memory System'' denotes the Mem0 pipeline (GPT-5-nano extraction + GPT-5-mini reader). LC = long-context, no memory extraction.}
\label{tab:accuracy-main}
\end{table*}

LC GPT-5-mini achieves the highest accuracy on LoCoMo and LongMemEval, exceeding the memory system by 35.2 and 33.4 percentage points respectively. On PersonaMem v2 the gap is smaller (7.3 percentage points), and the memory system marginally outperforms LC GPT-OSS-120B on that dataset.

The accuracy advantage of the long-context approach is expected: passing the full conversation history to the model preserves every detail, whereas the memory pipeline condenses conversations into a compact set of atomic facts (averaging 2,909 tokens per user after extraction from roughly 101,600-token conversations). Some information is inevitably lost during this compression step.

GPT-OSS-120B shows a mixed profile. On LoCoMo and PersonaMem v2 it scores substantially below GPT-5-mini (81.69\% vs.\ 92.85\% and 60.50\% vs.\ 69.75\%), while on LongMemEval its accuracy (48.20\%) is close to that of the memory system (49.00\%). These results suggest that model capability, not context length alone, is a limiting factor on the longer benchmarks.

% ── 4.2 Operational Cost ─────────────────────────────────────────────────────

\subsection{Operational Cost Breakdown}
\label{sec:cost-breakdown}

We measure the actual API costs incurred during the LongMemEval~\cite{wu2025longmemevalbenchmarkingchatassistants} experiments (500 conversations, mean context length 101,601 tokens) using the pricing described in \Cref{sec:cost-analysis}.

\paragraph{Memory System --- write phase.}
Fact extraction with GPT-5-nano across all 500 conversations cost \$21.76 in total (\$0.0435 per conversation), consuming 164.2M tokens (122.6M prompt, 41.6M completion). Embedding 103,183 extracted records (1.842M tokens) with \texttt{text-embedding-3-small} added a negligible \$0.037. The write phase is a one-time cost per user.

\paragraph{Memory System --- read phase.}
Answer generation from retrieved facts (top-$k = 20$, averaging 1,046 retrieved tokens per query) with GPT-5-mini cost \$0.65 across 500 evaluation questions, or roughly \$0.0013 per query.

\paragraph{Long-context baselines.}
Running GPT-5-mini on full conversations (\textgreater50k tokens) cost \$14.79 for 504 requests (\$0.0293 per request). GPT-OSS-120B cost \$7.65 for 664 requests (\$0.0115 per request) due to its lower per-token price. Both request counts exceed the 500 evaluation questions because transient API errors triggered automatic retries; each question is counted once toward accuracy but may have required more than one API call to complete.

In summary, the memory system front-loads cost at write time and then incurs a small, roughly fixed charge per query, whereas long-context LLMs pay a variable cost proportional to the full context on every turn.

% ── 4.3 Cost by Turn Count ───────────────────────────────────────────────────

\subsection{Cost Comparison by Interaction Volume}
\label{sec:cost-turns}

The cost trade-off between the two architectures changes as the number of interaction turns $N$ grows. \Cref{tab:cost-by-turn} shows the cumulative cost per user at a representative context length of 101,601 tokens (LongMemEval mean), using the cost model from \Cref{sec:cost-equations} with prompt caching applied to LC from turn 2 onward.

\begin{table*}[ht]
\centering
\begin{tabular}{r r r l}
\toprule
$N$ (turns) & \textbf{Memory System} & \textbf{LC GPT-5-mini} & \textbf{Result} \\
\midrule
1  & \$0.0450 & \$0.0265 & LC cheaper          \\
5  & \$0.0502 & \$0.0408 & LC cheaper          \\
10 & \$0.0568 & \$0.0588 & Memory cheaper      \\
15 & \$0.0634 & \$0.0768 & Memory saves $\approx$17\% \\
20 & \$0.0700 & \$0.0947 & Memory saves $\approx$26\% \\
\bottomrule
\end{tabular}
\caption{Cumulative cost per user at context length 101,601 tokens as a function of $N$ interaction turns. LC prices assume prompt caching from turn 2 onward (90\% discount on input tokens).}
\label{tab:cost-by-turn}
\end{table*}

For single-turn or low-turn interactions, the long-context approach is less expensive, because its first-turn cost (\$0.0265) is well below the memory system's write-plus-read cost (\$0.0450). Once the number of turns reaches approximately 10, the accumulated LC cost overtakes the memory system, which adds only \$0.0013 per additional turn. At $N = 20$, the memory system is 26\% cheaper.

This crossover arises from the structural difference between the two cost profiles: $C_\text{LC}(N)$ grows with $N$ even under caching (each cached turn still incurs an incremental charge proportional to context length), while $C_\text{Mem}(N)$ grows at the much smaller per-turn read rate.

% ── 4.4 Sensitivity Analysis ─────────────────────────────────────────────────

\subsection{Break-even Sensitivity Analysis}
\label{sec:sensitivity}

The break-even point $N_\text{BE}$, the turn count at which $C_\text{Mem} < C_\text{LC}$, depends on the conversation context length $L$. \Cref{tab:sensitivity} reports results for four representative context sizes.

\begin{table*}[ht]
\centering
\begin{tabular}{r r r r r}
\toprule
\textbf{Context $L$ (tokens)} & \textbf{Write Cost} & \textbf{LC Turn 1} & \textbf{LC Turn $N$} & \textbf{Break-even $N$} \\
\midrule
 30,000 & \$0.0129 & \$0.0086 & \$0.0018 & 13 \\
100,000 & \$0.0430 & \$0.0261 & \$0.0036 & 10 \\
200,000 & \$0.0859 & \$0.0511 & \$0.0061 &  9 \\
500,000 & \$0.2148 & \$0.1261 & \$0.0136 &  9 \\
\bottomrule
\end{tabular}
\caption{Break-even analysis across context sizes. Write Cost is the one-time memory extraction cost. LC costs are per-turn: Turn 1 reflects the uncached rate, Turn $N > 1$ reflects the cached rate. Break-even $N$ is the first turn at which the memory system becomes cheaper.}
\label{tab:sensitivity}
\end{table*}

Two observations follow from the table. First, the write cost grows proportionally with $L$ (fact extraction scales with input length), but the break-even $N$ decreases from 13 to 9 as $L$ increases from 30k to 500k tokens. This happens because the cached LC per-turn cost also grows with $L$, and it grows faster relative to the fixed read cost of the memory system. Second, the break-even $N$ is relatively insensitive to $L$ above 100k tokens: at 100k, 200k, and 500k, $N_\text{BE}$ differs by at most one turn.

\begin{figure*}[ht]
\centering
\includegraphics[width=0.95\linewidth]{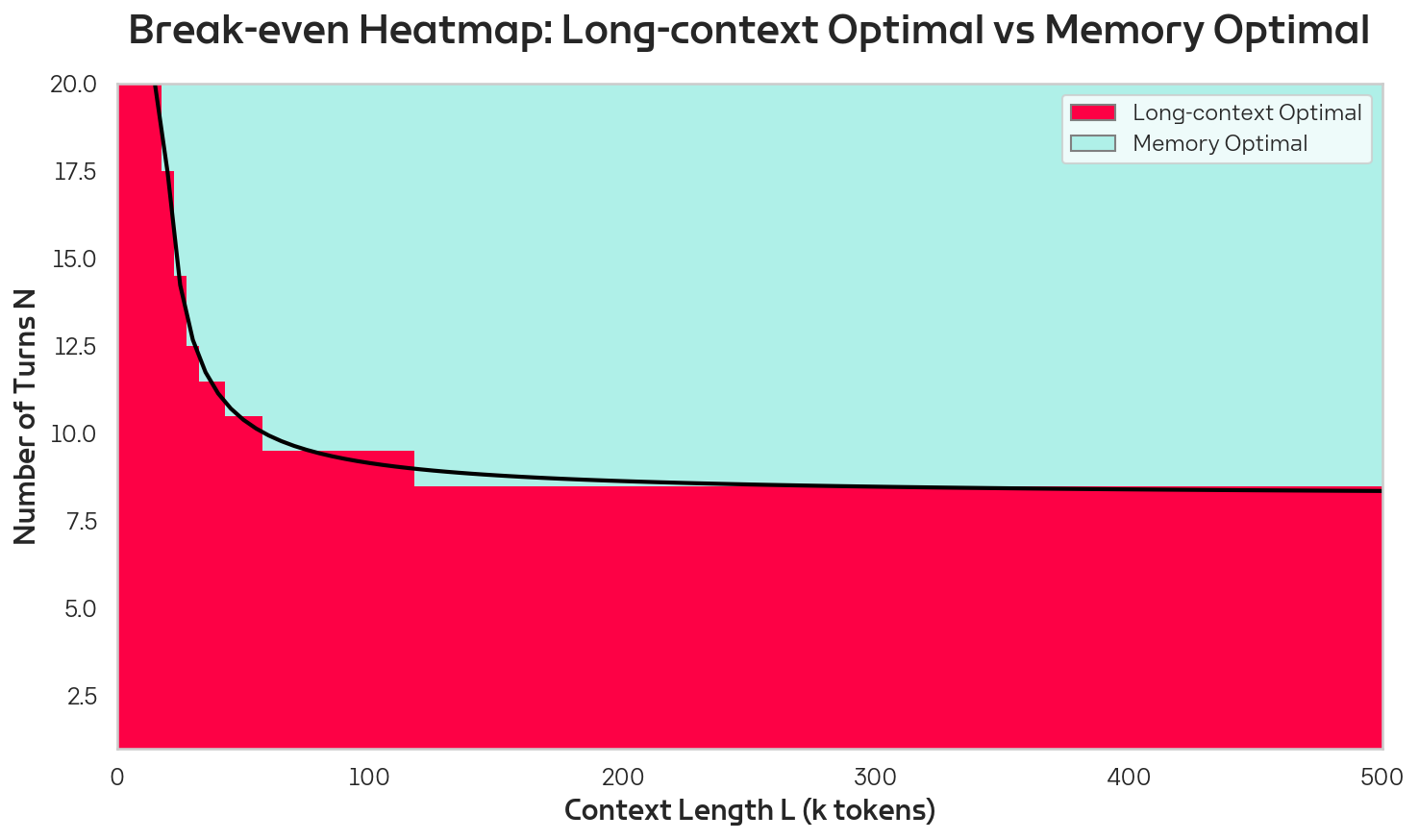}
\caption{Break-even heatmap: cumulative cost difference (LC minus Memory) as a function of context length $L$ and number of turns $N$. Red regions indicate where the long-context approach is cheaper; blue regions indicate where the memory system is cheaper. The black curve marks the break-even boundary.}
\label{fig:heatmap}
\end{figure*}

\Cref{fig:heatmap} visualizes the cost-optimal region across the full space of context lengths and turn counts. The boundary separating the two regions shifts leftward (lower $N$) as context grows, confirming that memory systems become cost-competitive more quickly in high-context settings.

\subsection{Summary}
\label{sec:summary}

The experimental results highlight a clear accuracy–cost trade-off between the two architectures. Long-context LLMs, particularly GPT-5-mini, achieve higher factual recall than the memory system across most benchmarks. However, this advantage comes at a cost that scales linearly with context length and interaction volume. For applications where a user interacts with the system more than approximately ten times — a reasonable expectation for any persistent assistant — the memory system becomes the more economical option, while remaining competitive in the subset of tasks where persona consistency matters most (PersonaMem v2~\cite{jiang2025personamemv2personalizedintelligencelearning}).

\section{Discussion}
\label{sec:discussion}

\subsection{Interpretation of Results}
\label{sec:interp}

The accuracy gap between the long-context approach and the memory system on LongMemEval and LoCoMo reflects a fundamental difference in how each architecture handles conversational history.
Passing the raw conversation to the model preserves everything the user said; the memory pipeline, by contrast, condenses each conversation into a compact set of atomic facts.
At a mean context of 101,601 tokens, extraction produces an average of 2,909 retrieved tokens per user—a compression ratio of roughly 35:1.
Under this compression, certain categories of information are irretrievably lost: precise temporal markers (\eg, ``she mentioned last Tuesday that\ldots''), implicit coreferences that are only resolvable by reading earlier turns, and ephemeral one-off details that a flat-typed extractor does not recognise as a stable fact.
\Cref{tab:loss-examples} illustrates the three failure modes with concrete examples drawn from the LongMemEval evaluation.
Questions that require multi-hop reasoning over any of these lost details will tend to be answered incorrectly.

\begin{table*}[ht]
\centering
\small
\begin{tabular}{>{\raggedright\arraybackslash}p{2.2cm} p{4.5cm} p{3.8cm} >{\raggedright\arraybackslash}p{2.2cm} >{\raggedright\arraybackslash}p{2.2cm}}
\toprule
\textbf{Loss type} & \textbf{Question} & \textbf{Retrieved memory (Mem0)} & \textbf{Mem0 answer} & \textbf{LC answer} \\
\midrule
Temporal marker &
  How many days passed between my visit to MoMA and the `Ancient Civilizations' exhibit at the Met? &
  ``Interested in art-themed day trips'' (visit dates not stored) &
  0 days & 7 days \\
\addlinespace
Implicit coreference &
  What was my previous occupation? &
  ``Previously managed a team of interns at a startup'' (activity stored, not the role title) &
  Managed interns at a startup & Marketing specialist at a startup \\
\addlinespace
\mbox{Ephemeral update} &
  How often do I attend yoga classes to help with my anxiety? &
  ``Yoga twice a week'' (earlier session; later update to three times not propagated) &
  Twice~a~week & Three times a week \\
\bottomrule
\end{tabular}
\caption{Concrete examples of information categories irretrievably lost under flat-typed fact extraction (LongMemEval). ``LC answer'' refers to LC GPT-5-mini, which reads the full conversation and answers correctly in all three cases.}
\label{tab:loss-examples}
\end{table*}

The smaller accuracy gap on PersonaMem v2 is consistent with the nature of that benchmark.
PersonaMem v2 tests whether a system maintains consistent responses about user-specific persona attributes~\cite{jiang2025personamemv2personalizedintelligencelearning}.
Such attributes—preferences, personal history, habitual behaviors—are precisely the kind of stable, factual properties that flat-typed memory extraction is designed to capture.
The memory system's narrow advantage over LC GPT-OSS-120B on this dataset (62.48\% vs.\ 60.50\%) suggests that explicit fact storage may offer benefits where the relevant information is distributable into discrete, addressable records.

The results for GPT-OSS-120B illustrate that model capability is a confounding factor.
On LoCoMo and PersonaMem v2, GPT-OSS-120B scores 10--15 percentage points below GPT-5-mini, despite receiving the same full-context input.
On LongMemEval, GPT-OSS-120B (48.20\%) is nearly identical to the memory system (49.00\%), but this convergence arises from GPT-OSS-120B underperforming on the longer, more complex histories, not from the memory system recovering its recall.
Context length alone does not determine accuracy; the capacity of the answer model to reason over long inputs also matters.

The cost results reinforce the argument made in the introduction.
Even with prompt caching at a 90\% discount on input tokens from the second turn onward, the per-turn cost of long-context inference grows linearly with context length, whereas the memory system's per-turn read cost remains roughly fixed at \$0.0013 per query.
This difference is independent of any particular pricing schedule and would persist under any pricing model that charges proportionally to input token count.

\subsection{Limitations}
\label{sec:limitations}

\paragraph{Single memory architecture.}
This study uses Mem0's flat-typed extraction pipeline as the sole memory baseline~\cite{chhikara2025mem0buildingproductionreadyai}.
More structured approaches—such as temporal semantic memory~\cite{su2026dialoguetimetemporalsemantic}, or hierarchical engram-based systems~\cite{hu2026evermemosselforganizingmemoryoperating}—may recover recall that flat extraction loses, and could alter the accuracy comparison.
The findings reported here characterize the accuracy–cost trade-off for flat-typed fact extraction specifically and should not be generalized to memory-augmented generation as a category.

\paragraph{Benchmark coverage.}
The three benchmarks cover distinct competencies but share a common focus on persistent user-specific information across multi-session dialogues.
They do not cover tasks that require real-time knowledge, reasoning over structured data, or compositional multi-document evidence chains.

\paragraph{Evaluation protocol.}
The LLM-as-a-judge framework using GPT-5-mini as evaluator~\cite{zheng2023judgingllmasajudgemtbenchchatbot} introduces its own sources of variance.
Although the 3-vote majority consensus protocol reduces stochastic fluctuation, the judge model may exhibit systematic biases toward responses generated by models in the same model family~\cite{zheng2023judgingllmasajudgemtbenchchatbot,ye2024justiceprejudicequantifyingbiases}, and may be more lenient on verbose near-misses than a human annotator would be~\cite{zheng2023judgingllmasajudgemtbenchchatbot,ye2024justiceprejudicequantifyingbiases,chen-etal-2024-humans}.

\paragraph{Cost model assumptions.}
The cost analysis is based on OpenAI API pricing as of the time of the experiments, with a fixed 90\% caching discount applied to input tokens from the second turn onward~\cite{openai2024promptcaching}.
Actual costs depend on cache hit rates, which vary with inter-request timing and prefix stability.
The model does not account for infrastructure costs (\eg, vector database hosting, embedding compute) beyond API charges, which would increase the absolute cost of the memory system but not change the qualitative shape of the break-even curve.
The cost comparison covers the GPT-5-mini pricing tier only; a corresponding breakdown for GPT-OSS-120B is excluded because its inference cost depends on the hosting provider and infrastructure, which differ across deployments.

\paragraph{Static memory snapshot.}
The benchmark protocol ingests all conversational data in a single write phase before any questions are posed.
This setup does not reflect deployments in which memory evolves during an active session—for instance, when a user adds new facts or revises prior statements while querying.
In practice, a memory system would update its fact store incrementally as new information arrives, which could shift the accuracy trade-offs reported here.
The effect of continuous memory updates on the cost--accuracy comparison is left to future work.

\subsection{Practical Implications}
\label{sec:implications}

The break-even analysis provides a concrete decision criterion for practitioners choosing between the two architectures.
At a context length of 100k tokens, the memory system becomes cheaper after approximately ten interaction turns (\Cref{sec:cost-turns,sec:sensitivity}).
For applications where a user is expected to re-engage with the same conversational context more than ten times—persistent personal assistants, long-running customer-support threads, multi-session tutoring systems—the memory system offers lower cumulative cost with a modest accuracy trade-off.
For single-session or low-repetition use cases, long-context inference is both cheaper at the first turn and more accurate overall.

The break-even point decreases as context length grows: at 500k tokens, the threshold drops to nine turns (\Cref{tab:sensitivity}).
Practitioners working with very long conversation histories therefore have a stronger economic incentive to adopt memory systems, even accounting for the write-phase cost.

The accuracy trade-off is task-dependent.
On benchmarks that test precise factual recall over complex histories (LongMemEval, LoCoMo), long-context models retain a material advantage at current extraction quality.
Where persona consistency is the primary requirement, the memory system is competitive.
Deployment decisions should take the query distribution of the target application into account when interpreting these results.

These findings have direct implications for enterprise AI deployments in which users interact repeatedly within the same conversational context—such as persistent data analytics assistants, long-running customer-support threads, or multi-session chatbots.
In such settings, the break-even criterion offers a concrete operational signal: when a user is expected to issue more than approximately ten queries against a 100k-token context, a memory-based architecture reduces cumulative inference cost while remaining competitive on tasks that require persona consistency, \eg, customer service chatbots, which need to maintain a professional and consistent brand alignment.
This trade-off is practically relevant because organizations increasingly deploy AI assistants that accumulate context over weeks or months, making the cost of processing the full conversation history on every query a significant operating expense.
Memory systems address this by replacing a growing per-query cost with a one-time extraction cost and a near-fixed per-query read cost, yielding more predictable, sub-linear cost scaling over time.
That being said, long-context inference remains preferable for workloads in which single-session precision is paramount; the break-even analysis should be treated as a principled decision criterion rather than a general recommendation.

\section{Conclusions and Future Work}
\label{sec:conclusion}

This paper compared a fact-based memory system (Mem0) against long-context LLM inference across three memory-centric benchmarks and a cost model incorporating prompt caching.
On accuracy, long-context GPT-5-mini significantly outperforms the memory system on LoCoMo and LongMemEval, while the memory system remains competitive on PersonaMem v2, where relevant information is well-suited to flat fact extraction.
On cost, the two architectures cross at approximately ten interaction turns at 100k tokens, with the break-even point decreasing as context length grows.
These findings suggest that the choice between memory systems and long-context inference is best treated as a deployment decision that depends on the nature of user queries, the number of interactions per user session, and the accuracy requirements of the application.
For deployments in which users issue more than approximately ten queries against the same conversational context, a memory-based architecture reduces cumulative inference cost while remaining competitive on persona-consistent tasks; long-context inference remains preferable for single-session or precision-critical workloads.

Future work should examine richer memory architectures—such as temporal semantic memory~\cite{su2026dialoguetimetemporalsemantic} and hierarchical engram-based systems~\cite{hu2026evermemosselforganizingmemoryoperating}—and extend the evaluation suite to cover tasks requiring real-time knowledge, structured-data reasoning, and compositional multi-document inference.

\subsection*{Acknowledgements}
\noindent
We would like to thank Anavit Jedmongkolwong, Kan Sirarojanakul, and Pornpailin Kobkiatkawin for their helpful comments, suggestions, and valuable discussions.

% \newpage
% \clearpage
\bibliography{custom}

% \newpage
% \clearpage

\onecolumn
\appendix
\crefalias{section}{appendix}

\section{Prompt Templates}
\label{sec:prompts}

\subsection{Fact Extraction: Memory Extraction Instructions}
\label{sec:prompts-extraction}

The following custom instructions are passed to the Mem0~\cite{chhikara2025mem0buildingproductionreadyai} \texttt{add} stage to guide memory extraction from raw conversations. These instructions were developed as part of the EverMemOS~\cite{hu2026evermemosselforganizingmemoryoperating} mem0 evaluation pipeline.

\begin{figure}[ht]
\begin{tcolorbox}[colback=gray!5, colframe=gray!40, title=Fact Extraction --- Custom Instructions]
\small
Generate personal memories that follow these guidelines:

\begin{enumerate}
  \item Each memory should be self-contained with complete context, including:
  \begin{itemize}
    \item The person's name (do not use ``user'' while creating memories)
    \item Personal details (career aspirations, hobbies, life circumstances)
    \item Emotional states and reactions
    \item Ongoing journeys or future plans
    \item Specific dates when events occurred
  \end{itemize}

  \item Include meaningful personal narratives focusing on:
  \begin{itemize}
    \item Identity and self-acceptance journeys
    \item Family planning and parenting
    \item Creative outlets and hobbies
    \item Mental health and self-care activities
    \item Career aspirations and education goals
    \item Important life events and milestones
  \end{itemize}

  \item Make each memory rich with specific details rather than general statements:
  \begin{itemize}
    \item Include timeframes (exact dates when possible)
    \item Name specific activities (e.g., ``charity race for mental health'' rather than just ``exercise'')
    \item Include emotional context and personal growth elements
  \end{itemize}

  \item Extract memories only from user messages, not incorporating assistant responses.

  \item Format each memory as a paragraph with a clear narrative structure that captures the person's experience, challenges, and aspirations.
\end{enumerate}
\end{tcolorbox}
\caption{Custom instructions passed to the Mem0 \texttt{add} stage for memory extraction.}
\label{fig:extraction-prompt}
\end{figure}

\subsection{LLM-as-a-Judge Grading Prompt}
\label{sec:prompts-judge}

The following system and user prompts are used by the GPT-5-mini~\cite{singh2025openaigpt5card} judge to evaluate generated answers against the golden answer~\cite{zheng2023judgingllmasajudgemtbenchchatbot}. This grading protocol was adopted from the EverMemOS~\cite{hu2026evermemosselforganizingmemoryoperating} evaluation framework.

\begin{figure}[ht]
\begin{tcolorbox}[colback=gray!5, colframe=gray!40, title=LLM-as-a-Judge --- System Prompt]
\small
You are an expert grader that determines if answers to questions match a gold standard answer.
\end{tcolorbox}
\vspace{4pt}
\begin{tcolorbox}[colback=gray!5, colframe=gray!40, title=LLM-as-a-Judge --- User Prompt Template]
\small
Your task is to label an answer to a question as `CORRECT' or `WRONG'. You will be given the following data: (1) a question (posed by one user to another user), (2) a `gold' (ground truth) answer, (3) a generated answer, which you will score as CORRECT/WRONG.

The point of the question is to ask about something one user should know about the other user based on their prior conversations. The gold answer will usually be a concise and short answer that includes the referenced topic. The generated answer might be much longer, but you should be generous with your grading---as long as it touches on the same topic as the gold answer, it should be counted as CORRECT.

For time-related questions, the gold answer will be a specific date, month, year, etc. The generated answer might be much longer or use relative time references (e.g., ``last Tuesday'' or ``next month''), but you should be generous with your grading---as long as it refers to the same date or time period as the gold answer, it should be counted as CORRECT. Even if the format differs (e.g., ``May 7th'' vs.\ ``7 May''), consider it CORRECT if it is the same date.

\textbf{Question:} \texttt{\{question\}}\\
\textbf{Gold answer:} \texttt{\{golden\_answer\}}\\
\textbf{Generated answer:} \texttt{\{generated\_answer\}}

First, provide a short (one sentence) explanation of your reasoning, then finish with CORRECT or WRONG. Do NOT include both CORRECT and WRONG in your response. Return the label in JSON format with the key \texttt{"label"}.
\end{tcolorbox}
\caption{System and user prompt templates used for LLM-as-a-judge evaluation.}
\label{fig:judge-prompt}
\end{figure}

\section{Model Parameters}
\label{sec:model-params}

\Cref{tab:model-params} lists the intended inference parameters for each model role. The GPT-5 model series (GPT-5-nano, GPT-5-mini) does not accept \texttt{temperature} or \texttt{max\_tokens} via the API and instead uses a \texttt{reasoning\_effort} parameter; for those models, the table reports reasoning effort. The GPT-OSS-120B model uses standard temperature and max-token settings.

\begin{table}[ht]
\centering
\begin{tabular}{l l l l r r}
\toprule
\textbf{Role} & \textbf{Model} & \textbf{Snapshot} & \textbf{Reasoning effort} & \textbf{Temp.} & \textbf{max\_tokens} \\
\midrule
Fact extractor   & GPT-5-nano   & \texttt{2025-08-07} & low    & — & — \\
Memory reader    & GPT-5-mini   & \texttt{2025-08-07} & medium & — & — \\
Long-context LLM & GPT-5-mini   & \texttt{2025-08-07} & low    & — & — \\
Long-context LLM & GPT-OSS-120B & —                   & —      & 0 & 2{,}000 \\
Judge            & GPT-5-mini   & \texttt{2025-08-07} & high   & — & — \\
\bottomrule
\end{tabular}
\caption{Intended model inference parameters. GPT-5-nano and GPT-5-mini use \texttt{reasoning\_effort} and are pinned to the \texttt{2025-08-07} snapshot; GPT-OSS-120B has no versioned snapshot and uses temperature and max\_tokens.}
\label{tab:model-params}
\end{table}

\section{Experimental Environment}
\label{sec:environment}

All LLM completions and embedding API calls were routed through OpenRouter\footnote{\url{https://openrouter.ai/api/v1}} as the unified API provider, which supports a broad range of model families under a single endpoint. Vector storage was provided by the pgvector\footnote{\url{https://github.com/pgvector/pgvector}} extension running on PostgreSQL~17 (Docker image \texttt{pgvector/pgvector:pg17}), deployed as a local service alongside the evaluation pipeline. Model pricing used in the cost-performance analysis is sourced from the OpenAI API pricing page.\footnote{\url{https://openai.com/api/pricing/}}

\end{document}